\documentclass[final]{cvpr}

\usepackage{times}
\usepackage{epsfig}
\usepackage{graphicx}
\usepackage{amsmath}
\usepackage{amssymb}
\usepackage{caption}
\usepackage{multirow}
\usepackage{subfigure}
\usepackage{booktabs}
\usepackage{wrapfig}
\usepackage{eucal}
\usepackage[pagebackref=false,breaklinks=true,letterpaper=true,colorlinks,bookmarks=false]{hyperref}

\newcommand{\lu}[1]{\textcolor{black}{#1}}

\newcommand{\HDMapGen}{HDMapGen\xspace}

\newcommand{\PAR}[1]{\vskip1pt \noindent {\bf #1~}}

\def\cvprPaperID{2979} % *** Enter the CVPR Paper ID here
\def\confYear{CVPR 2021}

\begin{document}

%%%%%%%%% TITLE
%\title{HDMapGen: High Definition Map Generation}
%\\with Autoregressive Hierarchical Graph Neural Networks}
\title{\HDMapGen: A Hierarchical Graph Generative Model of High Definition Maps}

% \author[1,2]{Lu Mi}
% % For a paper whose authors are all at the same institution,
% % omit the following lines up until the closing ``}''.
% % Additional authors and addresses can be added with ``\and'',
% % just like the second author.
% % To save space, use either the email address or home page, not both

% \author[1]{Hang Zhao}
% \author[3]{Charlie Nash}
% \author[1]{Xiaohan Jin}
% \author[1]{Jiyang Gao}
% \author[4]{Chen Sun}
% \author[4]{Cordelia Schmid}
% \author[2]{Nir Shavit}
% \author[1]{Yuning Chai}
% \author[1]{Dragomir Anguelov}

% \affil[1]{Waymo}
% \affil[2]{MIT CSAIL}
% \affil[3]{Tsinghua University}
% \affil[4]{DeepMind}
% \affil[5]{Google}
% %\affil[6]{Momenta.ai}

\author{
  \hspace{-1.3cm}
  \begin{tabular}[t]{c}
    Lu Mi$^{1,2,\dagger,*}$, Hang Zhao$^{1,*}$, Charlie Nash$^3$, Xiaohan Jin$^1$, Jiyang Gao$^1$,  Chen Sun$^4$, \\ Cordelia Schmid$^4$, Nir Shavit$^2$, Yuning Chai$^1$, Dragomir Anguelov$^1$\\
    $^1$Waymo, $^2$MIT, $^3$DeepMind, $^4$Google\\
\end{tabular}
}

\maketitle
\pagestyle{empty}
\thispagestyle{empty}

\let\thefootnote\relax\footnotetext{\leftline{$\dagger$Work done during internship at Waymo. }}

\let\thefootnote\relax\footnotetext{\leftline{$^*$Corresponding to: \texttt{lumi@mit.edu}, \texttt{zhaohang0124@gmail.com}}}

% \twocolumn[{
% \renewcommand\twocolumn[1][]{#1}
% \maketitle
% \begin{center}
%     \centering
% \includegraphics[width=1\linewidth]{./figures/paper/teaser.png}
% \captionof{figure}{Hierarchical graph output from \HDMapGen.}
% \label{fig:intro_results_teaser}
% \end{center}
% \vspace{1em}
% }]

%%%%%%%%% ABSTRACT
\begin{abstract}

High Definition (HD) maps are maps with precise definitions of road lanes with rich semantics of the traffic rules. They are critical for several key stages in an autonomous driving system, including motion forecasting and planning. However, there are only a small amount of real-world road topologies and geometries, which significantly limits our ability to test out the self-driving stack to generalize onto new unseen scenarios. To address this issue, we introduce a new challenging task to generate HD maps. 
In this work, we explore several autoregressive models using different data representations, including sequence, plain graph, and hierarchical graph. We propose \HDMapGen, a hierarchical graph generation model capable of producing high-quality and diverse HD maps through a coarse-to-fine approach. Experiments on the Argoverse dataset and an in-house dataset show that \HDMapGen significantly outperforms baseline methods. Additionally, we demonstrate that \HDMapGen achieves high scalability and efficiency. 

\end{abstract}
\section{Introduction}

\begin{figure}[h]
\captionsetup{font=small}
\begin{center}
%\vspace{-40pt}
\includegraphics[width=1\linewidth]{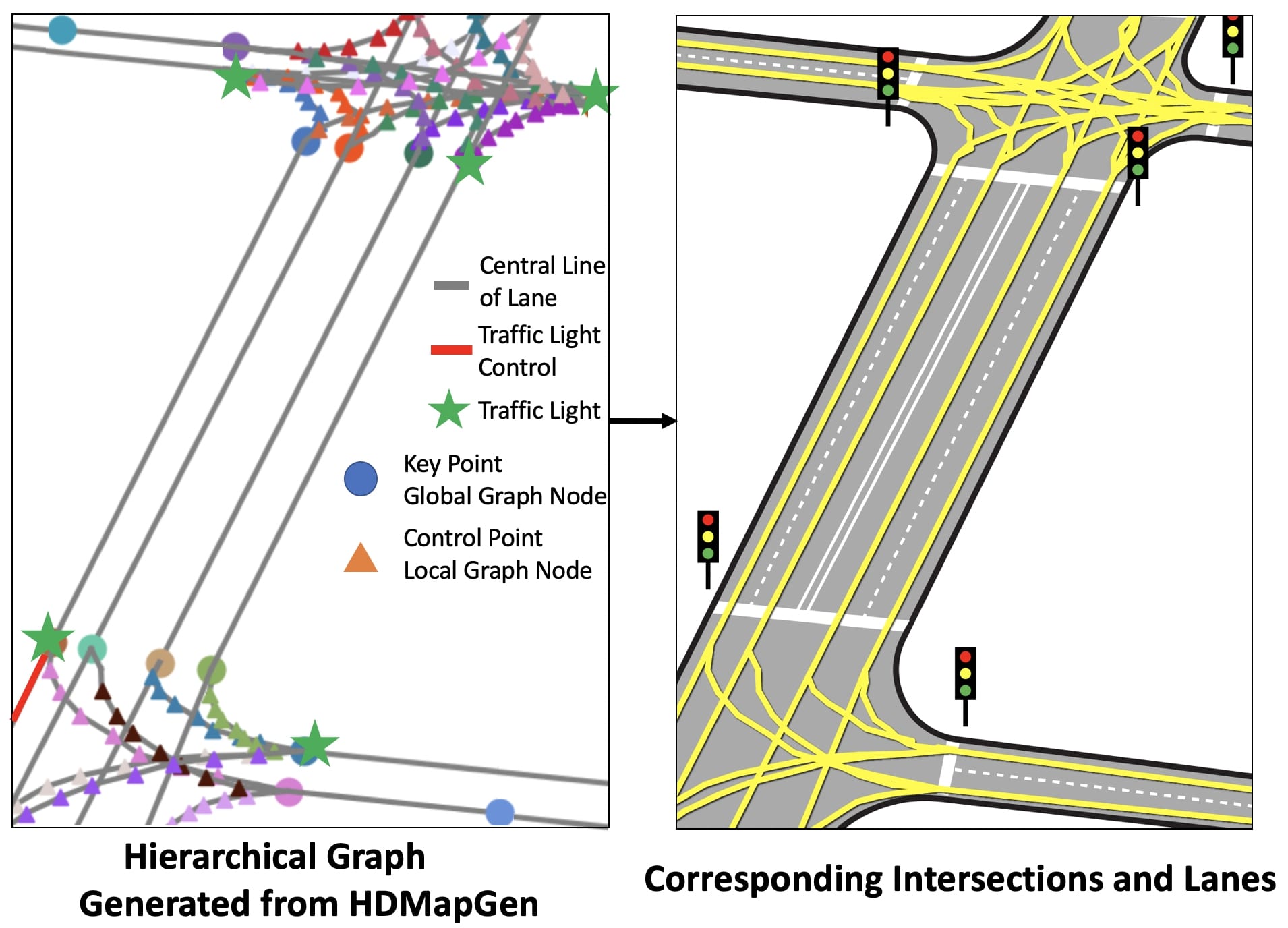}
\end{center}
\vspace{-0.4cm}
\caption{Left: a hierarchical map graph generated from \HDMapGen. One traffic light controlled lane is shown as an example; Right: a map with corresponding intersections and lanes rendered from the hierarchical map graph. \vspace{-0.4cm}}
\label{fig: teaser}
%\vskip -0.3cm
\end{figure}

% Definition of HD maps
High Definition maps (HD maps) are electronic maps with precise depictions of the physical roads, usually with an accuracy of centimeters, together with rich semantics of traffic rules, such as one-way-street, stop, yield, \etc.
% Importance and applications
HD maps sit at the core of autonomous driving applications as they provide a strong prior for the self-driving robots to localize themselves in the 3D space~\cite{bauer2016using,thrun2007simultaneous}, predict other vehicles' motions~\cite{chai2019multipath,zhao2020tnt} and maneuver themselves around~\cite{chen2010high,lavalle1998rapidly}.
Furthermore, HD maps are critical building blocks of city modeling and simulation. Simulating novel city environments finds a wide range of applications in game design and urban planning.

%Furthermore, HD maps are the building blocks of driving simulators. To simulate a realistic environment for robots drive in, engineers often define and build the abstract HD maps of a scene first, and then perform graphical rendering. 
%Beyond autonomous driving, HD maps also find applications in gaming and urban planning.

% Challenge
Practically, HD maps are constructed according to a strict mapping procedure: first, a fleet of vehicles with mapping sensor suites (containing LiDAR, Radar, and camera) are sent to capture the scene; then, the sensor data are processed and stitched together to obtain map imagery; finally, human specialists annotate on top of the imagery to provide vectorized representations of the world geometry with semantic attributes.
On the other hand, we aim to produce synthetic HD maps in a data-driven way, mainly due to two reasons: (1) building HD maps from the real world is prohibitively expensive; (2) the small number of real-world maps prevent us from testing the generalization capability of self-driving stacks in simulation, such as motion forecasting and motion planning.
% Due to the high cost, building a large number of lane maps from the real world is prohibitive. 
%However, the demand of HD maps is increasing: urban designers rely on city simulations to structure their designs; autonomous driving simulations require a humongous number of environments and scenarios to train a well-performing robot agent.
% Therefore, an effective map generation method would be a remedy for the data scarcity problems in autonomous driving simulation and urban design.
% In this paper, we focus on lane maps of HD maps, which define the lane geometry and their semantic attributes.

% Existing works
Existing methods in modeling cities and maps are mostly relying on procedural modeling, and hand-crafted generation rules~\cite{parish2001procedural}, and are therefore not flexible and adaptable to new scenarios.
There is no previous attempt to generate HD maps using modern deep generative models to the best of our knowledge.
The most related works are those which generate city layouts~\cite{chu2019neural}. Compared to HD maps, city layouts are not suitable for autonomous driving applications since they only contain coarse locations of the roads (with a resolution of roughly 10 meters) and lack details such as lanes of the roads or traffic lights.

% Unique problems we need to address.
Unique attributes of HD maps pose new challenges to modeling them: (1) HD maps are composed of physical road elements with geometric features, \eg marked straight lanes and hypothesized turning lanes; (2) Lane maps contain rich semantic attributes, \eg the direction of lanes, the association between traffic lights and lanes. By addressing these challenges, our major contributions in this work are as follows:

\begin{itemize}
\item We pose a new important and challenging problem to generate HD lane maps in a data-driven way.
% \item We propose a hierarchical graph generative model which largely outperforms other baselines. It is capable of generating high-quality HD maps with sufficient feasibility and diversity.
\item We perform a systematic exploration of modern autoregressive generative models with different data representations and propose \HDMapGen, a hierarchical graph generative model that largely outperforms other baselines. 
\item We evaluate our model on the maps of the public Argoverse dataset and an in-house dataset, covering cities of Miami, Pittsburg, and San Francisco. Results show that our model produces maps with high fidelity, diversity, scalability, and efficiency.

\end{itemize}
\section{Related Work}

\subsection{Street Map Modeling and Generation}
City street modeling and generation is an important component of computer-aided urban design. Most classical works rely on procedural modeling methods~\cite{cityengine, aliaga2008interactive,watson2008procedural,benes2014procedural, tensorstreet2008chen, emilien2015worldbrush, galin2011hierarchical, nishida2015example, parish2001procedural, yang2013urban}. One of the more well-known methods is the L-system~\cite{parish2001procedural}, which generates road networks from a sequence of instructions defined by hand-crafted production rules. 
To impose more control over the generated results, later works proposed to sample from procedural models according to constraints or likelihood functions~\cite{talton2011talton}.
However, these methods are still constrained by the rules, therefore are not flexible in terms of generating maps with different city styles.

In recent years, deep learning methods have been applied to map reconstruction and encoding.
Several works~\cite{bastani2018roadtracer,Homayounfar_2019_ICCV, li2018polymapper, mattyus2017deeproadmapper} attempted to extract and reconstruct road topologies from overhead images.
VectorNet~\cite{gao2020vectornet} and LaneGCN~\cite{liang2020learning} proposed to efficiently encode roads using graph attention networks and graph convolutions, replacing traditional rendering-based models.
Chu \etal propose Neural Turtle Graphics (NTG)~\cite{chu2019neural}, a generative model to produce roads iteratively. They use an encoder-decoder RNN model that encodes incoming paths into a node and decodes outgoing nodes and edges. The goal of NTG is to generate city-level road layouts. In comparison, we focus on high definition
%($2\times$ orders of resolution) 
road and lane generation, which is much more challenging.

\subsection{Graph Generative Models}
Our model architecture is inspired by the rapid progress in the area of graph generation spearheaded by seminal works such as graph recurrent neural networks (GRNN)~\cite{hajiramezanali2019variational}, graph generative adversarial networks (GraphGAN)~\cite{wang2017graphgan}, variational graph auto-encoder (VGAE)~\cite{kipf2016variational}, and graph recurrent attention networks (GRAN)~\cite{liao2019efficient}.
More recently, Cao~\etal proposed MolGAN~\cite{de2018molgan}, and Samanta~\etal proposed NEVAE~\cite{samanta2020nevae} to generate small molecule graphs.
These methods all focus on generating graph topologies, such as adjacency matrices. In comparison, HD lane maps are graphs that come with spatial coordinates and geometric features, which poses another layer of complexity to the generation problem.

\subsection{Geometric Data Generation}
Given the geometric nature of maps, another line of related work is geometric data generation.
To generate 2D sketch drawings, Ha~\etal proposed SketchRNN~\cite{ha2017neural}, an RNN model with a VAE structure to sequentially produce sketch strokes following human drawing sequences. To assemble furniture from primitive parts, GRASS~\cite{li2017grass} and StructureNet~\cite{mo2019structurenet} adopted several variations of auto-encoding models. More recently, Nash~\etal proposed PolyGen~\cite{nash2020polygen}, an autoregressive transformer model that generates 3D furniture meshes. Compared to human sketches, furniture, and molecules, graphs of lane maps usually consist of more nodes and edges. Instead of treating the whole map as a sequence, our method generates a map with a two-level hierarchy, greatly improving the generation quality.

\section{Method}
In Section~\ref{subsec:Data Representations}, we explore different data representations to generate HD maps  and demonstrate the efficiency of using hierarchical graph. In Section~\ref{subsec:Autoregressive Modeling}, we introduce our autoregressive graph generative model \HDMapGen that uses a hierarchical graph as data representation.

%gran is not we proposed

\begin{figure*}[t]
\captionsetup{font=small}
\begin{center}
%\vspace{-40pt}
\vskip -0.6cm
\includegraphics[width=0.9\linewidth]{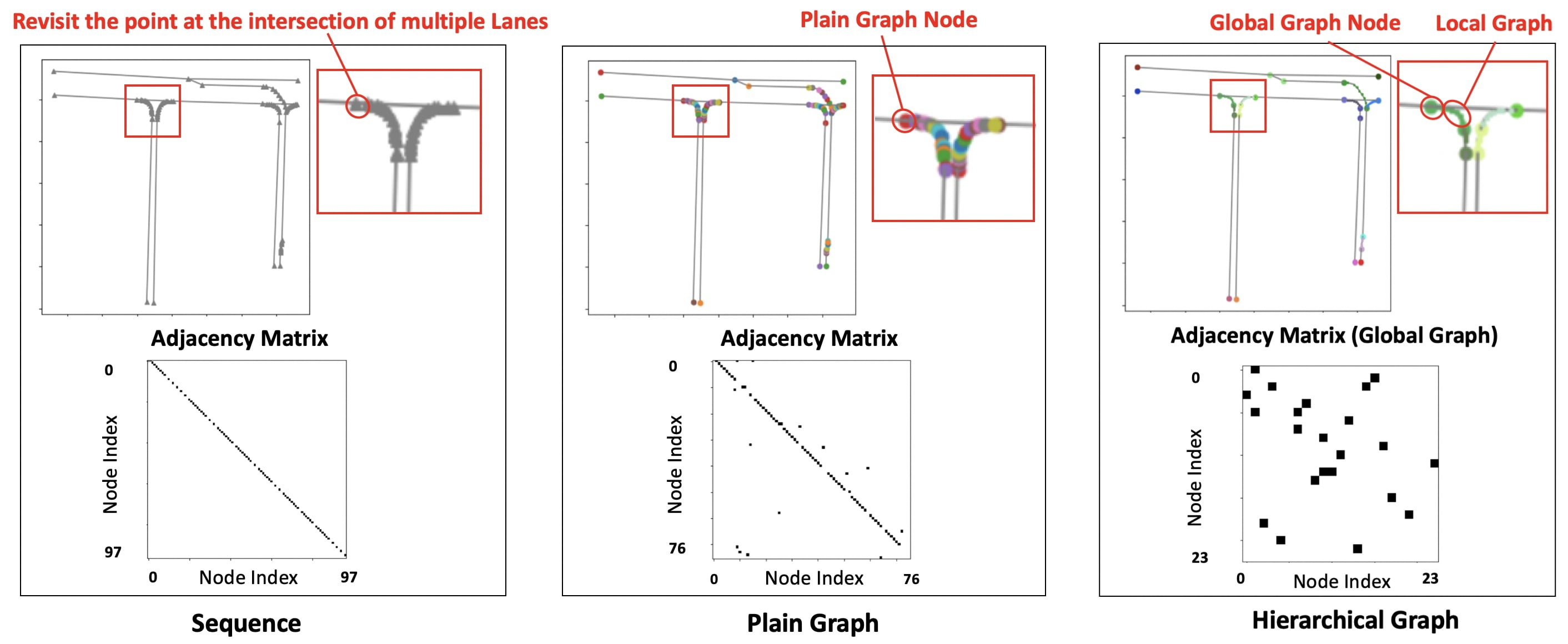}
\end{center}
\vskip -0.6cm
\caption{Different data representations of HD maps. Sequence representation leads to a big diagonal adjacency matrix, and it also suffers from revisiting the intersection points multiple times; plain graph representation results in a big sparse adjacency matrix; our proposed hierarchical graph representation gives a smaller and denser adjacency matrix (of the global graph), which reduces generation difficulty and improves efficiency.\vspace{-0.4cm}}
\label{fig: data_representation}
%\vskip -0.3cm
\end{figure*}

\subsection{Data Representation}
\label{subsec:Data Representations}

To provide detailed road information to the autonomous vehicle, HD maps used in applications such as autonomous driving typically contain many essential components, including lanes, boundaries, crosswalks, traffic lights, stop signs, \etc.
In this work, we focus on generating lanes, which are a core component of HD maps.
A lane is usually represented geometrically by its central line as a curve, and the curve is stored as a polyline with a sufficient amount of control points on it to reconstruct the curvature. Our goal is to generate these control points of central lane lines.
Meanwhile, each lane has a unique ID. Its predecessor and successor lane IDs are also provided. Moreover, the lanes also have semantic attributes such as whether any traffic light controls it or not. 

For lane map generation, there are various ways to represent the lane objects. We explore three different data representations: a sequence, a plain graph, and a hierarchical graph, as shown in Figure~\ref{fig: data_representation}.
We will later show that the hierarchical graph representation that \HDMapGen utilizes is the most efficient and scalable. Moreover, it vastly outperforms other representation methods.

\noindent\textbf{Sequence.}
While assuming all lines (lanes in our case) are connected, a straightforward representation is to sort all the lines in a map and treat the entire map as one sequence of points. This approach was adopted by SketchRNN~\cite{ha2017neural}. Each point in the sequence has an offset distance in the $x$ and $y$ direction  $(\Delta x, \Delta y)$ from the previous point, and a state variable $q \in \{1, 2, 3\}$. The state $q = 1$ indicates that the next point is starting a new lane object; the state $q = 2$ identifies the next point to continue in the same lane object, and the state $q = 3$ indicates that the entire map generation has completed.
One of the major disadvantages of this method is that there is no perfect way to pre-define a sequential ordering of these points. Secondly, to cover an intersection point of multiple lanes requires revisiting that point more than once. Under this representation, an intersection point will be represented as multiple independent points in the sequence.

\begin{figure*}[t]
\captionsetup{font=small}
\begin{center}
\vspace{-0.7cm}
\includegraphics[width=0.9\linewidth]{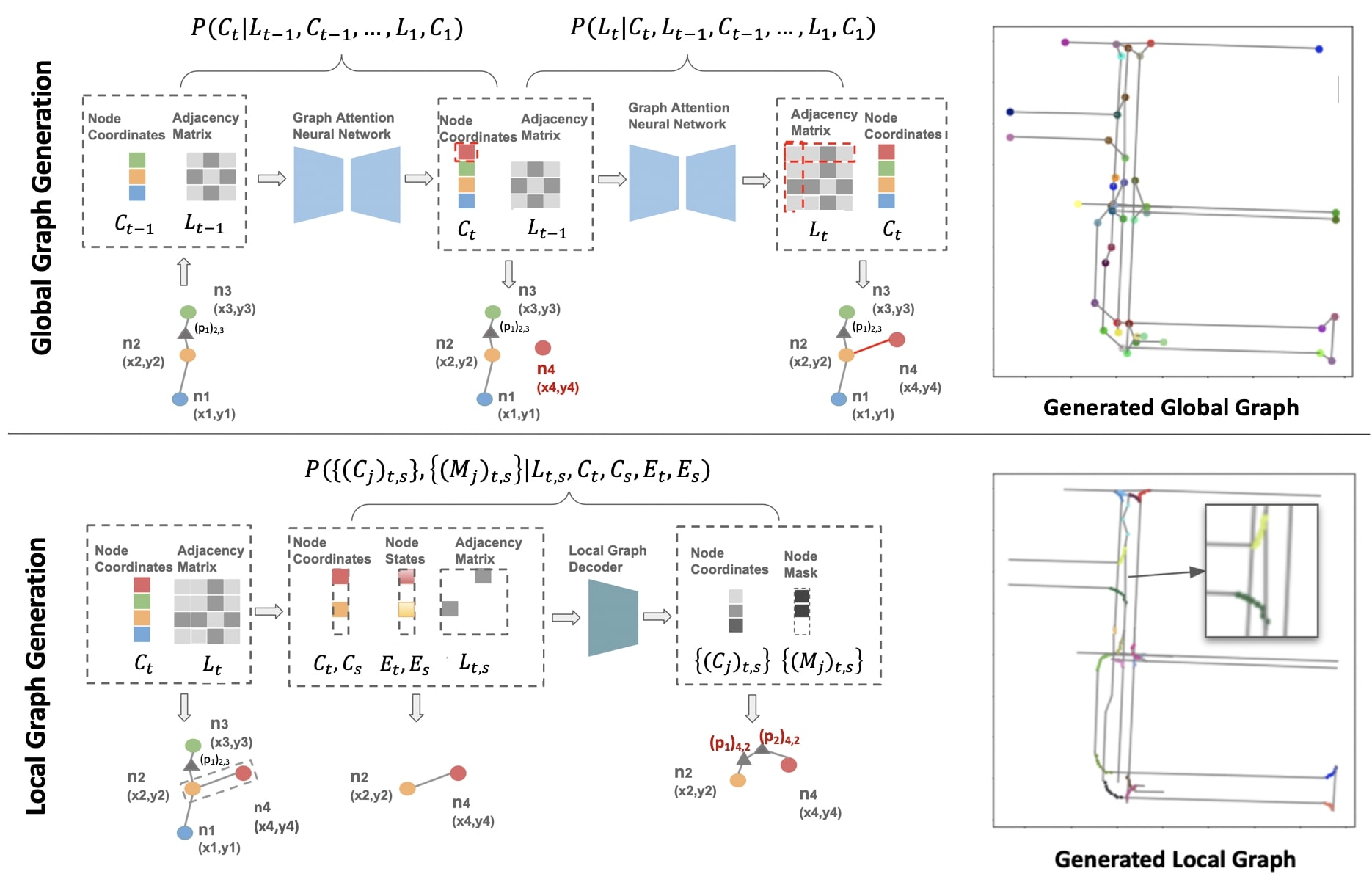}
\end{center}
\vspace{-0.6cm}
\caption{\HDMapGen pipeline. At each step, global graph generator (top) produces a new node with its coordinates and its connections to existing nodes, which is a row and a column in the adjacency matrix; local graph generator (bottom) further decodes coordinates of local nodes between the two connected global nodes. We also demonstrate the global graph and the local graph generated from \HDMapGen. \vspace{-0.3cm}}
\label{fig:method_description}
%\vskip -0.3cm
\end{figure*}

\noindent\textbf{Plain Graph.} The second baseline is to use a plain graph to represent a map. This strategy was adopted in recent work on road layout generation NTG~\cite{chu2019neural}. Under this representation, a plain graph contains all control points as nodes and all line connections between control points as edges. Moreover, each node has a node attribute of 2D coordinates $(x, y)$.
%and each line connection of two points is an edge between two nodes in the graph. 
The points inside a lane object have a degree of two, while the intersection points have a degree greater than two. This representation solves the revisiting issue of sequence generation. However, the method is still inefficient due to a large number of control points to guarantee a high-resolution map. 
To generate an edge between every pair of control points, it resorts to a very large adjacency matrix. 

\noindent\textbf{Hierarchical Graph (\HDMapGen).} Here, we propose an efficient and scalable alternative to use a hierarchical graph to represent the map.
Under this approach, we first construct a \textit{global graph} with \textit{key points} as its nodes. Key points are defined as the endpoints of lanes or the intersection points of multiple lanes. 
Then the edge of the global graph represents whether a lane exists between two key points. Since key points are a small subset ($30\%$) of original points,
which require a much smaller adjacency matrix. %Given a map containing $N$ lane objects, the global graph edges could be predicted with a complexity of $O(N^2)$.
In the second step, we represent each lane's curvatures details with a \textit{local graph}. Each lane is constructed with a uniquely defined sequence of \textit{control points} between two key points. Given this fixed topology, we could directly predict the coordinates of the local nodes. Moreover, we also predict a corresponding node mask to allow variations of the number of control points during generation.
%we then design a single-shot decoder of the local graph, the output is a sequence of local node coordinates and corresponding node mask to allow variations of the number of control points in each lane during generation.
%we could significantly reduce the complexity to predict all the edges from $O(N^2M^2)$ to $O(N^2 + NM)$.
As demonstrated in Figure~\ref{fig: data_representation}, using a hierarchical graph has a better performance to preserve the natural hierarchical structure information of the HD map. Moreover, it also enables a smaller size of the adjacency matrix and higher efficiency than using plain graph and sequence. 

\subsection{Autoregressive Modeling}
\label{subsec:Autoregressive Modeling}
Autoregressive modeling has achieved remarkable performance in many generation tasks\cite{gregor2014deep,nash2020polygen}, especially for sequential data such as speech~\cite{oord2016wavenet} and text~\cite{sutskever2014sequence}. 
More recently, several studies such as GraphRNN~\cite{you2018graphrnn} and GRAN~\cite{liao2019efficient} also took an autoregressive approach for the graph generation. In this work, we also use autoregressive modeling for HD map generation. 

%\textcolor{red}{We incorporate the GRAN model to generate the global topology}
%We adapt Graph Attention Neural Network (GRAN) as the basic graph neural network structure for our model, according to the state-of-the-art efficiency and sample quality of this model. 
%One of the main differences between our approach and GRAN is that GRAN is designed for plain graph generation, while we build a hierarchical graph generative model by combining a global graph decoder and a local graph decoder. Meanwhile, the spatial graph we tackle has additional node attributes, while GRAN and GraphRNN are only used to generate a graph's topology. In our model, we include a geometry Multilayer Perceptron (MLP) encoder and decoder for coordinates. 
%also demonstrate a better performance than these ``one shot'' methods such as GraphVAE []. Thus we also follow their strategy to apply autoregressive modeling to our problem. 
%We propose a novel graph generative model that not only generates topological structures (represented by adjacency matrix $L$), but also spatial coordinates of the nodes $C$. This is achieved by an additional geometry encoder and decoder to produce node attributes. 

 Our proposed \HDMapGen is a two-level hierarchical graph generative model, as shown in Figure~\ref{fig:method_description}.  Firstly, the global graph generation process outputs a graph with both topological structures represented by adjacency matrix $L$, and the geometric features represented by nodes' spatial coordinates $C$. Each node of the global graph is generated autoregressively. In our task, we first model the global graph as an undirected graph and later assign the lane direction information as the sequential order in the local graph. At the current step $t$, the global graph decoder predicts the $C_{t}$ of this new node at $t$ and all corresponding edges (lanes) $L_{t,s}$ between previously generated nodes $s$ in a single shot, where $s \in [1,t-1]$. Then, it constructs a new row $L_t$ and column (due to the symmetry for the undirected graph) of the adjacency matrix $L$. Then, we apply with a local graph decoder, which outputs the curvature details in each lane object $L_{t,s}$. The local graph outputs for each global edge $L_{t,s}$ are two sequences with a fixed length $W$ of coordinates $\{(C_j)_{t,s}\}$ and valid mask $\{(M_j)_{t,s}\}$ for each local node $j$, where $j \in [1, W]$. The mask defines which node is valid to allow a variation of the number of nodes in each local graph. The local graph decoder is also generating the local graphs for all lane objects $L_{t,s}$ in a single shot, which enables very fast and efficient running. We further add another single-shot semantic attribute decoder to predict semantic features of all generated edges $L_{t,s}$. We introduce the details of the key components of \HDMapGen in the following text.

\PAR{Global Graph Generation.} 
To apply autoregressive modeling, we firstly factorize the probability of generating adjacency matrix $L$ and coordinates $C$ of the global graph as
\begin{align*}
P(L,C)=\prod_{t=1}^{T} P(L_{t},C_{t} \mid L_{1}, C_{1}, \cdots, L_{t-1},C_{t-1}),
\end{align*}
where $P(L_{t},C_{t} \mid L_{1}, C_{1}, \cdots, L_{t-1},C_{t-1})$ defines that at the current step $t$, the conditional probability of generating every new node's attributes $C_{t}$ and its edges $L_{t}$ with nodes generated from all previous steps $1$ to $t-1$. 

Furthermore, since the process of generating $L_{t}$ and $C_{t}$ at each step $t$ are not independent, we explore three variants of global graph decoder to study the performance with different priority of generating $L_{t}$ or $C_{t}$. We use \textit{coordinate-first} to refer to generating node coordinate first and then topology, defined as $P(L_{t},C_{t}) = P(L_{t}\mid C_{t})P(C_{t})$, as shown in Figure~\ref{fig:method_description}. And we use \textit{topology-first} to refer to first generating graph topology and then generate node coordinates, defined as $P(L_{t},C_{t}) = P(C_{t}\mid L_{t})P(L_{t})$. Another simplified strategy ignores the dependency between nodes and edges is named as \textit{independent}, defined as $P(L_{t},C_{t}) = P(C_{t})P(L_{t})$.
More details are in Supplementary.

Inspired by GRAN \cite{liao2019efficient}, which has state-of-the-art performance in sample quality and time efficiency, we also incorporate a recurrent graph attention network for the global graph generation. The graph attention module \cite{velivckovic2017graph} performs the node state update with the attentive message passing, enabling a better representation of global information. 
%\lu{The global graph generation uses graph attention network \cite{liao2019efficient} encoding the spatial graph inputs. This network using attentive messages for message passing. 
In the model, we firstly define an initial node state $E_{s}^0$ before the message passing at each step $t$ as
$E_{s}^0 = W_L L_s + W_C C_s + b, s \in [1,t-1]$,
where $W_L$, $W_C$ and $b$ are parameters of the MLP encoders taking topology and geometry inputs. Then for all nodes including new node $t$ and nodes $s$ already generated, we perform multiple runs of message passing. For each run $r$, the message $m_{i,k}^r$ between node $i$ and its neighbor node $k \in \EuScript{N}(i)$ is $m_{i,k}^r = f(E_i^r - E_k^r)$, where $\EuScript{N}(i)$ are the neighbor nodes of $i$. And a binary mask is defined as $B_i=0$ to indicate if node $i$ is already generated, or defined as $B_i=1$ if under construction at step $t$. The node state $E_i^r$ is further concatenated with this mask $B_i$ into $\Tilde{E_i^r} = [E_i^r,B_i]$. And the attention weights $a_{ik}^r$ for edge $L_{i,k}$ is defined as $ a_{ik}^r = Sigmoid( g(\Tilde{E_i^r} - \Tilde{E_i^k}))$.  And the node state update is then calculated as $E_i^{r+1} = GRU(E_i^{r},\sum
_{k \in \EuScript{N}(i)}a_{ik}^rm_{ik}^r)$. In this experiment, we use MLPs for both $f$ and $g$. The GNN model has $7$ layers and a propagation number of $1$. After the message passing is completed, we take the final node states $E_t$ and $E_s$ after message passing, and use model $MLP(E_t - E_s)$ to decode them into a mixture of Bernoulli distribution for topology outputs $L_{t}$. And another model $MLP(E_t)$ is applied to generate a 2D Gaussian mixture model (GMM) for coordinate outputs $C_{t}$.
%To apply the graph attention network, we define an initial node state $E_{i}^0$ before the message passing at each step $t$ as
%\begin{align*}
%E_{i}^0 = W_L L_i + W_C C_i + b, \forall i < t,
%\end{align*}
%where $W_L$, $W_C$ and $b$ are parameters of the MLP encoders taking topology and coordinates inputs. At each step $t$, we use a mixture of Bernoulli distributions to model the topology output $P(L_{t})$ \cite{liao2019efficient}. And we use a 2D Gaussian mixture model (GMM) with normal distributions for coordinate output $P(C_{t})$,
%and 
The entire model is optimized with the minimization of negative log-likelihood (NLL) for coordinate outputs and binary cross-entropy (BCE) for topology outputs. And for the GMM, we further add a temperature term $\tau$ to control the diversity (variance) during the sampling. The original standard deviation $\sigma_x$ and $\sigma_y$ in GMM are modified into $\sigma_x \tau$ and $\sigma_y \tau$. We follow the common teacher-forcing strategy for autoregressive model training by providing ground truth of $L_{1}, C_{1}, \cdots, L_{t-1},C_{t-1}$ as inputs when predicting $P(L_{t},C_{t})$, which avoids performing the reparameterization trick for the sampling process during the backpropagation. 

\PAR{Local Graph Generation.} 
%At the step $t$, after a new global graph node and its corresponding edges with previous nodes, it is generated. Then for the edge between global node $t$ and previously generated node $s$, a local graph will be further generated. 
%where $E_{t}$ is the state of global node $t$ after message passing. $E_{t}$ contains the global graph embedding to infer $C_{t}$ and $L_{t}$.
\lu{As the local graph to represent the curvature details of lanes always has a unique topology as a sequence of control points, we use a padding vector with a fixed
maximum length $W$ to represent the sequence. We model it as a sequence of coordinate output $\{(C_j)_{t,s}\}$ and a corresponding valid mask $\{(M_j)_{t,s}\}$, where $j \in [1, W]$. Then the conditional probability of the local graph during generation is defined as $P(\{(C_j)_{t,s}\},\{(M_j)_{t,s}\} \mid L_{t,s}, C_{t}, C_{s}, E_{t}, E_{s})$. The valid mask enables a variation in the number of nodes in each graph. For the straight lines which have been filtered with redundant control points after map preprocessing, the valid mask $\{(M_j)_{t,s}\}$ is all $0s$, while for the lanes at the corner which usually have multiple control points remained to guarantee the smoothness, $\{(M_j)_{t,s}\}$ is likely to have more values of $1$. And we use an MLP model to generate the local graph. The models are optimized with the minimization of mean square error (MSE) for coordinate output $\{(C_j)_{t,s}\}$, and with the minimization of BCE for the valid mask $\{(M_j)_{t,s}\}$.}

\PAR{Semantic Attribute Generation.} Like the local graph generation, we predict an additional attribute, traffic lights feature for the generated edge. Given a lane $L_{t,s}$ between node $t$ and node $s$, we predict whether the lane is controlled by traffic lights %by providing $E_t$ and $E_s$ as input to 
with an additional edge feature decoder using MLP. We train this edge feature decoder for binary classification with the minimization of BCE.   

%More details about the autoregressive modeling are described in Supplementary.

%\subsection{Graph Attention Neural %Network}
%\label{subsec:Graph Attention Neural Network}

\section{Experiments}
In this section, we demonstrate the efficacy of our proposed \HDMapGen model to generate high-quality maps. We explore three autoregressive generative models for sequence, plain graph, and hierarchical graph data representations and evaluate generation quality, diversity, scalability, and efficiency.

\subsection{Dataset and Implementation Details}

\PAR{Datasets.} We evaluate the performance of our models on two datasets across three cities in the United States. One is the public Argoverse dataset \cite{chang2019argoverse}, which covers Miami (total size of 204 kilometers of lanes) and Pittsburg (total size of 86 kilometers). We randomly sample 12000 maps with a field-of-view (FoV) of $200m \times 200m$ as the training dataset for Miami and 5000 maps with the same FoV for Pittsburg. We also evaluate an in-house dataset, which covers maps from the city of San Francisco. We train on $6000$ maps with a FoV of $120m\times120m$.

\PAR{Map Preprocessing.} %from Vectorized Data to Hierarchical Graph}
The key components of HD maps, the central line of drivable lanes, are usually over-sampled to guarantee a sufficient resolution. However, graph neural network is limited in scalability. So in a pre-processing step, we remove redundant control points with a small variation of curvature in each lane. This step enables us to remove $70\%$ of points while not compromise the map quality. We then define a hierarchical spatial graph based on the pre-processed vector map. We use Depth-First-Search (DFS) to construct the global graph's adjacency matrix and define the generation order for autoregressive modeling. \lu{The start node is randomly selected.} Then we define the sequence of control points between global nodes as a local graph. The sequential order is the same as the lane direction. 

\PAR{Implementation Details.}\HDMapGen uses graph attention neural network as the core part to perform the attentive message passing for graph inputs. We use multi-layer-perceptrons (MLPs) for all other encoding and decoding steps. The graph node coordinates are normalized to $[-1,1]$. The model is trained on a single Tesla V100 GPU with the Adam optimizer, where we use an initial learning rate of 0.0001 with a momentum of 0.9. 

\subsection{Baselines}
% \subsubsection{Baselines on Sequence and Plain Graph} 
%Graph Generative Model}
We compare the quality of our hierarchical graph generative model \HDMapGen with a sequence generative model in SketchRNN \cite{ha2017neural} and a plain graph generative model PlainGen derived from the global generation step of \HDMapGen.

\PAR{SketchRNN.} We map all control points of the lane objects inside each HD map into a sequence for a sequence generative model. Each sequential data point has a pair of 2D coordinates and a state variable to define each line's continuity. We choose an ascending order of spatial coordinates to define the sequential order. We explore
conditional and unconditional variants for this model. For the conditional generation, a target map is provided as both input and target as a variational auto-encoder during training. Meanwhile, a target map is still provided as input during sampling. For an unconditional generation, a decoder-only model is trained to generate the target maps.

\PAR{PlainGen.}
The plain graph generative model has the same implementation as the global graph generative model in \HDMapGen, and it takes the entire plain graph as inputs. Every control point in the map is taken as a global node in the plain graph. A corresponding edge is constructed to define whether these two nodes are connected. We use DFS to construct the adjacency matrix and define the order of generation. And we explore two variants of \textit{coordinate-first} and \textit{topology-first} for this model.

%The second baseline we have is to apply plain graph generative model. We adopt the same global graph generative model in HDMapGen to perform the generation on the defined plain graph. 

\subsection{Qualitative Analysis}

\begin{figure}[t]
\captionsetup{font=small}
\begin{center}
%\vspace{-40pt}
 \vspace{-0.2cm}
\includegraphics[width=1\linewidth]{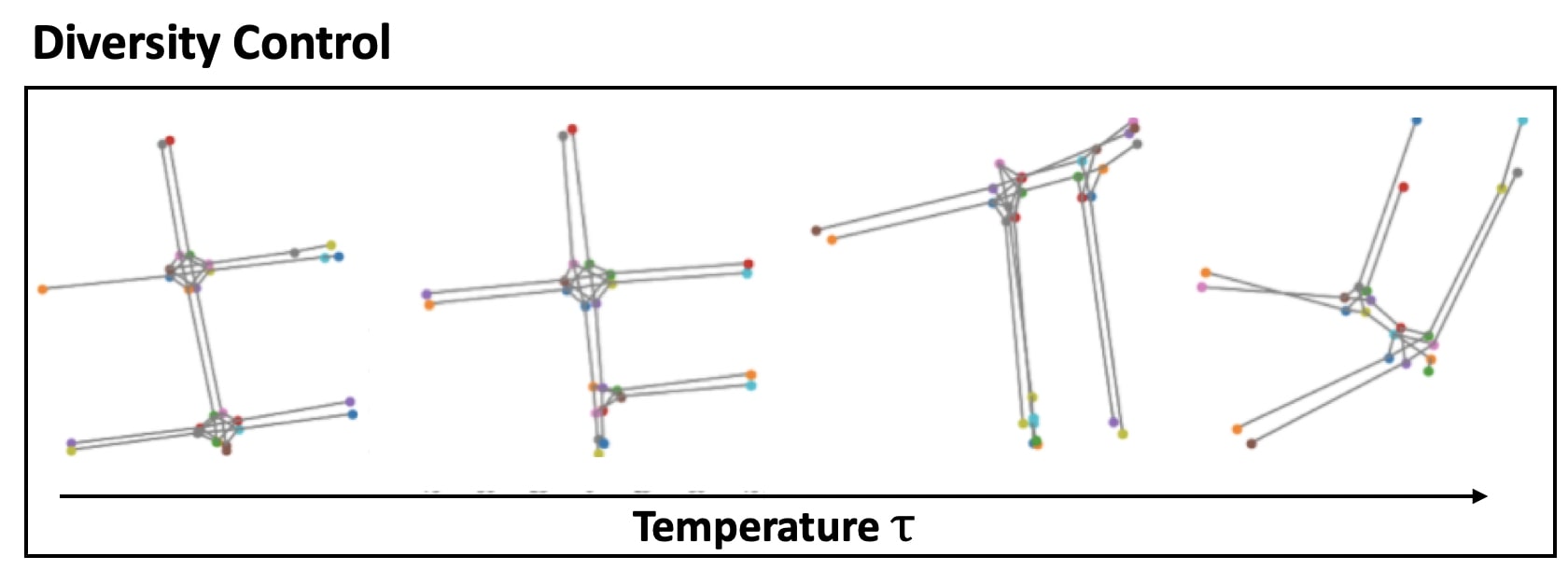}
\end{center}
 \vspace{-0.5cm}
\caption{The diversity of the global graph from HDMapGen is improved as temperature $\tau$ increases, but the realism suffers.\vspace{-0.3cm}}
%The results are from HDMapGen model trained on the in-house dataset.}
\label{fig: global_graph_generation}
%\vskip -0.3cm
\end{figure}

\PAR{Global Graph Generation.} We first show the global graph from \HDMapGen in Figure~\ref{fig: global_graph_generation}. We could see with only a small set of global nodes could represent a typical pattern in HD maps to include two parallel crossroads. We also demonstrate that the diversity of model outputs could be improved by increasing the temperature $\tau$. While the diversity and realism trade-off still exist during generation, we find an empirical bound of 0.2 for $\tau$ to guarantee high-quality samples in our experiments.

\begin{figure}[t]
\captionsetup{font=small}
\begin{center}
%\vspace{-0.7cm}
\includegraphics[width=0.8\linewidth]{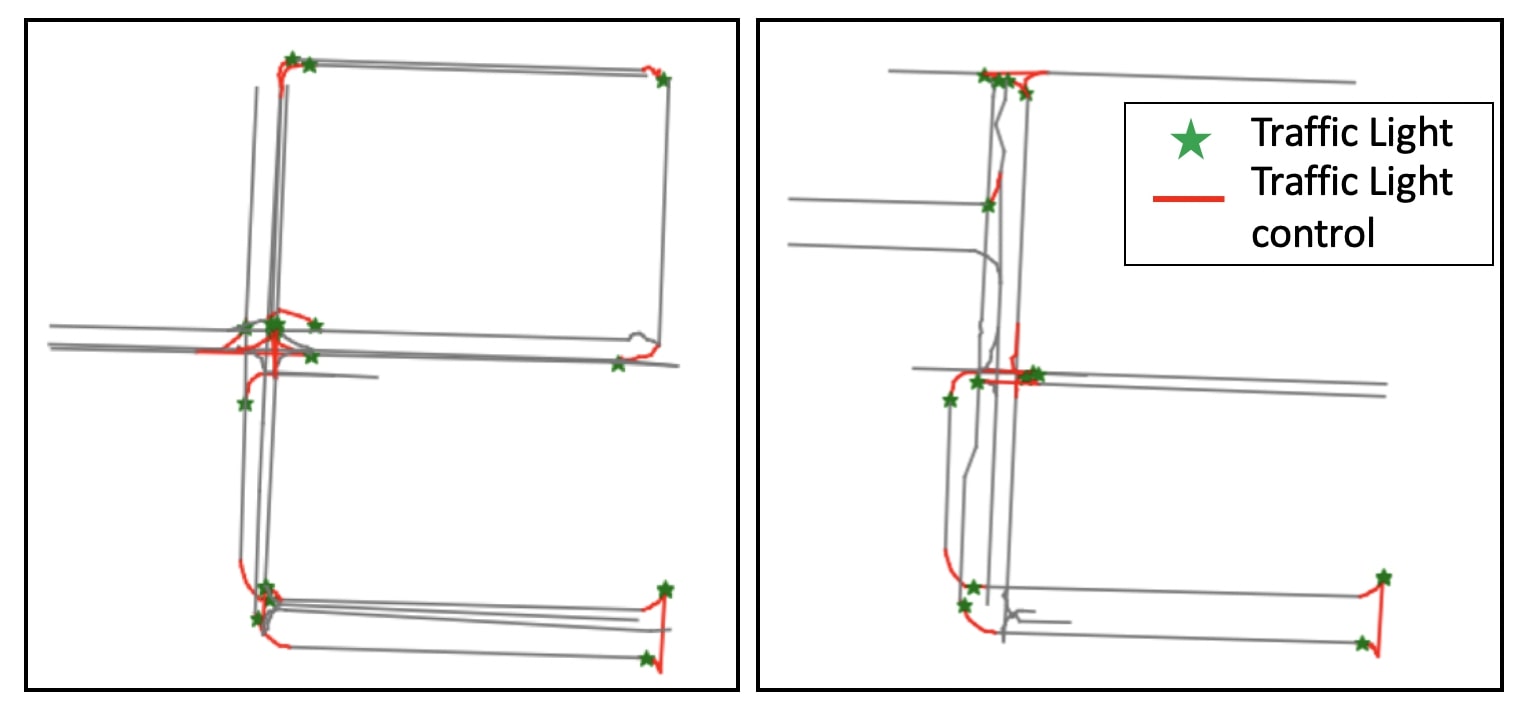}
 \end{center}
 \vspace{-0.5cm}
\caption{Semantic attributes of traffic lights and local graph of lanes controlled by traffic light generated from \HDMapGen.\vspace{-0.4cm}}
 \label{fig: traffic_light}
 %\vskip -0.3cm
 \end{figure}

\PAR{Local Graph Generation.} %The local graph is generated after each step of a node and corresponding edges are constructed in global graph generation. 
The local graph represents a sequence of control points that reveal the curvature details of each lane object. For lanes, the density of control points usually increases at the corner of each lane. In Figure~\ref{fig: traffic_light}, we show our model is capable of generating such features to have smooth curves at the corner of each lane.

\PAR{Semantic Attribute Generation.} As shown in  Figure~\ref{fig: traffic_light}, the traffic lights and the traffic light controlled lanes generated from \HDMapGen are mostly located on the crossroads and turn roads. The generated semantic attributes are consistent with the real-world urban scenes. 

 \begin{figure*}[t]
\captionsetup{font=small}
\begin{center}
\vspace{-1cm}
\includegraphics[width=0.98\linewidth]{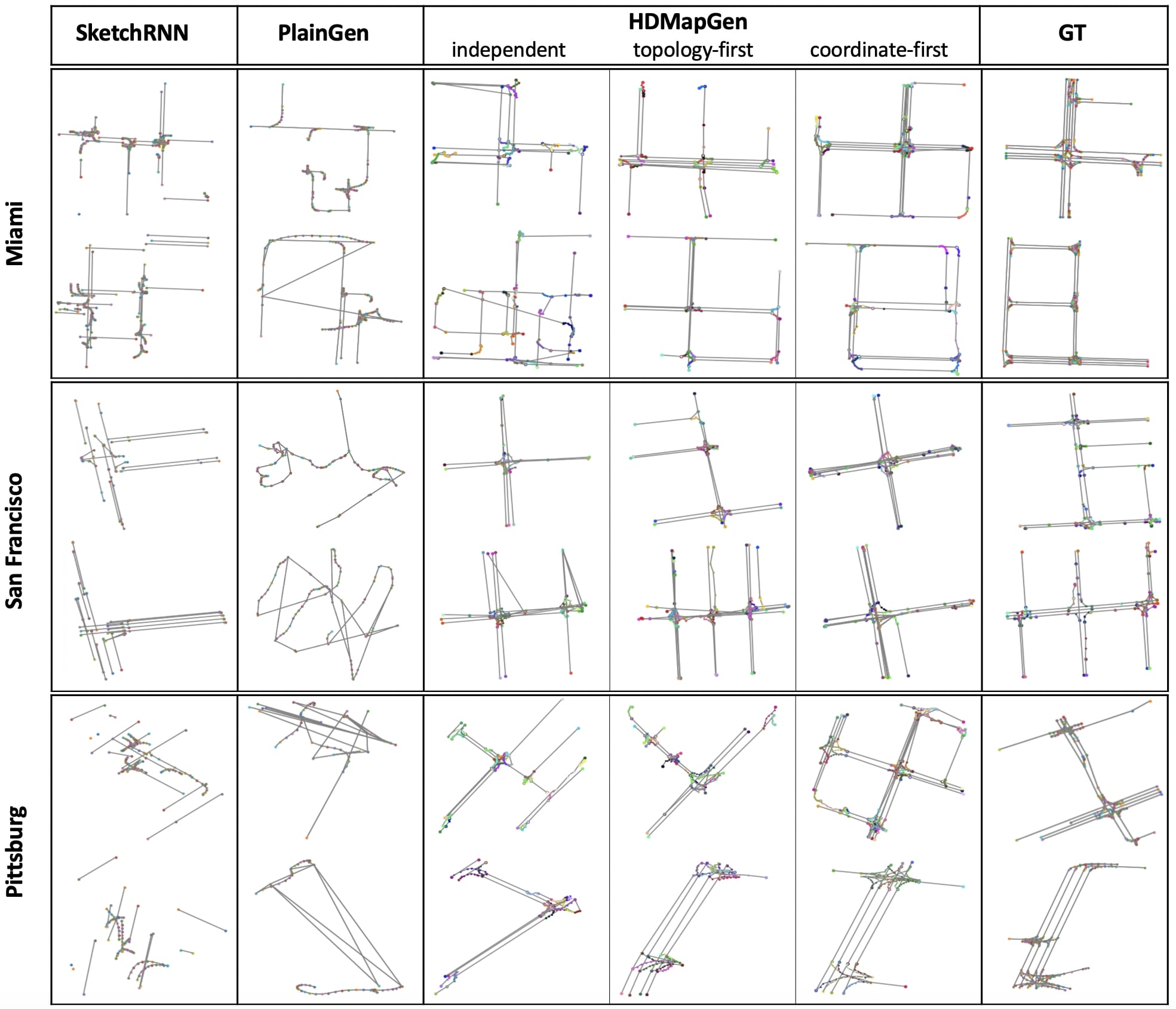}
\end{center}
\vskip -0.5cm
\caption{HD Maps with different city styles from hierarchical graph generative models HDMapGen (\textit{coordinate-first}, \textit{topology-first} and \textit{independent}) outperform plain graph generative model PlainGen, and sequence generative model sketchRNN.}
\vskip -0.4cm
\label{fig: result_baseline}
\end{figure*}

\PAR{Comparison with Baselines.} In Figure~\ref{fig: result_baseline}, we show the qualitative comparison between our models and other baselines. The results show that the proposed hierarchical graph generative model HDMapGen (both \textit{coordinate-first} and \textit{topology-first}) generate the highest quality maps and vastly outperform other methods. They capture the typical features of HD maps, including patterns like the overall layouts, the crossroads,  parallel lanes, \etc. Moreover, they are capable of generating maps with different city styles. For the maps generated from HDMapGen with the \textit{independent} model, the problematic crossing of lanes happens frequently, as the model fails to take the dependence of coordinates and topology into account. For PlainGen, we find the model to completely fail for this task, as it is too challenging to generate a graph with a much large number of nodes and edges in this setting, compared to the hierarchical graph. For SketchRNN, we find the model can learn the overall geometric patterns. However, there are a large number of problematic cut-offs or dead-ends occurring in the generated lanes. This is because, during the sequential data generation, it is possible to stop generating a consecutive point at any step and then re-start at any arbitrary location. This property is not an issue for a sketch drawing task. However, for HD map generation, the model needs to guarantee the continuity constraint between generated points. 

\subsection{Quantitative Analysis}

\subsubsection{Metrics}

We design four metrics to quantify the generation results:
% including the topology fidelity, the geometry fidelity, and a set of urban planning statistics.

\PAR{Topology fidelity.} We use maximum mean discrepancy (MMD) \cite{hajiramezanali2019variational} to quantify the similarity with real maps on graph statistics using Gaussian kernels with the first Wasserstein distance. The topology statistics we apply here is the \textit{degree} distributions and \textit{spectrum} of graph Laplacian.

\PAR{Geometry fidelity.} We use geometry features, the \textit{length}, and \textit{orientation} of lanes to quantify the similarity with real maps. Then we compute the Fréchet Distance to measure two normal distributions using these features. 
%Such statistics is able to measure the geometry similarity with the target.

\PAR{Urban planning.} We use the common urban planning features, \textit{connectivity} to represent node degree, node \textit{density} within a region, \textit{reach} as the number of accessible lane within a region to evaluate the transportation plausibility, \lu{and \textit{convenience} as the Dijkstra shortest path length for node pairs of our generated maps compared with real maps} \cite{chu2019neural}. We also use Fréchet Distance to measure two normal distributions of these features. 

\begin{table*}[t]
\captionsetup{font=small}
\begin{center}
\vspace{-0.5cm}
\begin{tabular}{lc|cccc|cc|cc} 
%\hline
\toprule
\multicolumn{2}{c|}{Measurement}&\multicolumn{4}{c|}{Urban Planning} &\multicolumn{2}{c|}{Geometry Fidelity} &\multicolumn{2}{c}{Topology Fidelity}\\
\hline
\multicolumn{2}{c|}{Metrics}&\multicolumn{6}{c|}{Fréchet Distance} &\multicolumn{2}{c}{MMD}\\
\hline
 \multicolumn{2}{c|}{Features} &Conne. &Densi. &Reach. &Conve. &Len. &Orien. &Deg. &Spec. \\ \cline{1-2}
\multicolumn{2}{c|}{Methods} &$10^{0}$ &$10^{1}$ &$10^{1}$ &$10^{1}$ &$10^{-1}$ &$10^{1}$ &$10^{-1}$ &$10^{-1}$\\
\hline
 
 \multirow{2}{*}{SketchRNN} &conditional &0.50 &21.20 &13.43 &49.4 &2.04 &1.77 &1.03  &4.94 \\ 
  &unconditional &0.44 &41.12 &29.83 &49.5 &1.64 &1.22 &0.83  &4.74  \\\cline{1-2}
  
 \multirow{2}{*}{PlainGen}  
 &topology-first &0.54 &5.18 &14.57 &22.6 &1.85 &1.54 &0.17   &0.31  \\ 
 &coordinate-first &0.39 &4.30 &4.81 &12.0 &1.64 &0.46 &0.12  &\textbf{0.29}  \\\cline{1-2}
 
 \multirow{3}{*}{HDMapGen} &independent &0.31 &4.38 &\textbf{4.22} &11.2 &\textbf{1.49} &0.52 &0.06 &0.54  \\ 
  &topology-first &0.26 &\textbf{4.06} &4.47 &\textbf{9.9} &\textbf{1.49} &\textbf{0.37} &\textbf{0.05} &0.63 \\ 
  &coordinate-first &\textbf{0.17} &4.79 &4.74 &11.8 &\textbf{1.49} &0.53 &0.07  &0.90 \\ 
\bottomrule
%\hline
\end{tabular}
\end{center}
\vspace{-0.5cm}
\caption{Measurements of urban planning, geometry fidelity and topology fidelity on \HDMapGen and baselines PlainGen and SketchRNN. Urban planing (features of \textit{connectivity}, \textit{density}, \textit{reach} and \textit{convenience}) and Geometry (features of the \textit{length} and \textit{orientation} of lanes) are quantified with a Fréchet Distance metric.  Topology fidelity (features of node \textit{degree} and \textit{spectrum} of graph Laplacian) is quantified with a MMD metric. For all metrics, lower is better. Results are evaluated on Argoverse  dataset. \vspace{-0.3cm}}
\label{Tab: Quantitative Evaluation}
\end{table*}

%\textbf{Chamfer like distance}: we also use the chamfer like distance to quantify the map-wise similarity of each generative models, and finally provide a measurement of diversity.

As shown in Table~\ref{Tab: Quantitative Evaluation}, \HDMapGen (both \textit{coordinate-first} and \textit{topology-first}) outperform the baselines on most of the metrics. The results are consistent with the qualitative analysis: SketchRNN has a poor performance in topology-related features, such as the degree. Since problematic cut-offs or dead-ends of lanes frequently happen in SketchRNN, so the generated maps have a very different distribution of node degrees from real maps. PlainGen has a better performance in the spectrum of graph Laplacian. As we evaluate topology features by transforming outputs of all models into the plain graph level, which might preserve the intrinsic topology patterns for outputs from PlainGen.

\begin{table}[h]
\captionsetup{font=small}
\small
\setlength{\tabcolsep}{6pt}
%\vspace{-0.2cm}
\begin{center}
\begin{tabular}{ c c| c c c c c}
\toprule
\multicolumn{2}{c|}{Temperature} & 0.1 & 0.2  & 0.3  & 0.4  & 0.5 \\
\hline
Diversity &GT &11.6 &11.9 &11.1 &11.6 &\textbf{13.0}\\ 
Diversity &Output &2.75 &4.83 &5.83 &6.62 &\textbf{8.34} \\ 
\toprule

\end{tabular}
\end{center}
\vspace{-0.6cm}
\caption{Diversity quantified by Chamfer distance (scaling by $10^4$) for global graph generated from \HDMapGen using different temperatures $\tau$. Novelty is compared to ground truth or among outputs internally. Results are evaluated on our in-house dataset.}
\vspace{-0.4cm}
\label{tab: diversity: chamfer}
\end{table}

\PAR{Diversity.} \lu{We use a metric quantified by Chamfer distance to quantify the map-wise diversity of the global graph generated from HDMapGen. As shown in Table~\ref{tab: diversity: chamfer}, we evaluate the diversity on two levels, one is the novelty compared to real map ground truth, the other is the internal diversity among output samples.
It shows that results generated with varied temperatures are novel compared with real maps.
%, especially for the greedy sampling (the temperature of $0$), which has the largest Chamfer distance. Since it's likely to generate an over-simplified while still plausible pattern, which is different from the ground truth maps. 
For internal diversity, we could see the Chamfer distance drastically increases as the temperature becomes larger}.
%, which is consistent with the qualitative results.}

\subsection{Ablation Studies}

We further conduct ablation studies on the impact of using different dependence and generation priority on three variants of \HDMapGen models \textit{coordinate-first}, \textit{topology-first} and \textit{independence}.

% We evaluate the generation quality and diversity from our hierarchical graph generative model \HDMapGen with three variants of \textit{coordinate-first}, \textit{topology-first} and \textit{independent} with  different  definitions  of  priority  and  dependence on coordinates and topology generation.

We show the ablation results in Table~\ref{tab: ablation_studies}. For \textit{coordinate-first} model, BCE, which quantifies the topology prediction performance, is the best optimized. As after coordinates are generated as a prior, then the topology prediction could be better optimized. Meanwhile, for \textit{topology-first} model, \lu{NLL, which quantifies the geometry prediction performance, is the best optimized with the generated topology as prior.} In contrast, \textit{independence} model is the worst as it fails to model the dependence of coordinate and topology during generation. 

% \begin{table}[h]
% \captionsetup{font=small}
% \small
% \begin{center}
% \begin{tabular}{ c |c c c|c c c}
%  \toprule
%  Model &\multicolumn{3}{|c}{HDMapGen} &\multicolumn{3}{|c}{PlainGen}\\
%  \hline
%  Metrics & \textit{cof.} & \textit{tof.}  & \textit{ind.} & \textit{cof.} & \textit{tof.}  & \textit{ind.}\\ 
%  \hline
%  NLL & -5.61  & \textbf{-6.14}  & -4.49 &-16.83 &-23.48 &\\  
%  BCE &\textbf{0.001}  &0.050  & 0.053 &0.001 &0.013 &\\  
% \bottomrule
% %\hline
% \end{tabular}
% \end{center}
% \vspace{-0.4cm}
% \caption{Negative Log likelihood (NLL) and binary cross entropy (BCE) on Argoverse dataset for HDMapGen and PlainGen with three variants as \textit{coordinate-first}, \textit{topology-first} and \textit{independent}. \vspace{-0.4cm}}
% \label{tab: ablation_studies}
% \end{table}

\begin{table}[h]
\captionsetup{font=small}
\small
\vspace{-0.2cm}
\begin{center}
\begin{tabular}{ c c c c}
 \toprule
 Metrics & coordinate-first & topology-first  & independent\\ 
 \hline
 NLL & -5.610  & \textbf{-6.142}  & -4.490\\  
 BCE &\textbf{0.001}  &0.050  & 0.053 \\  
\bottomrule
%\hline
\end{tabular}
\end{center}
\vspace{-0.5cm}
\caption{\lu{Negative Log likelihood (NLL) and binary cross-entropy (BCE) on the generated global graph from three variants of HDMapGen including \textit{coordinate-first}, \textit{topology-first} and \textit{independent}. Results are evaluated on Argoverse dataset. \vspace{-0.5cm}}}
\label{tab: ablation_studies}
\end{table}

\subsection{Scalability Analysis}

\begin{figure}
\captionsetup{font=small}
\begin{center}
%\vskip -0.5cm
\vspace{-0.2cm}
\includegraphics[width=0.9\linewidth]{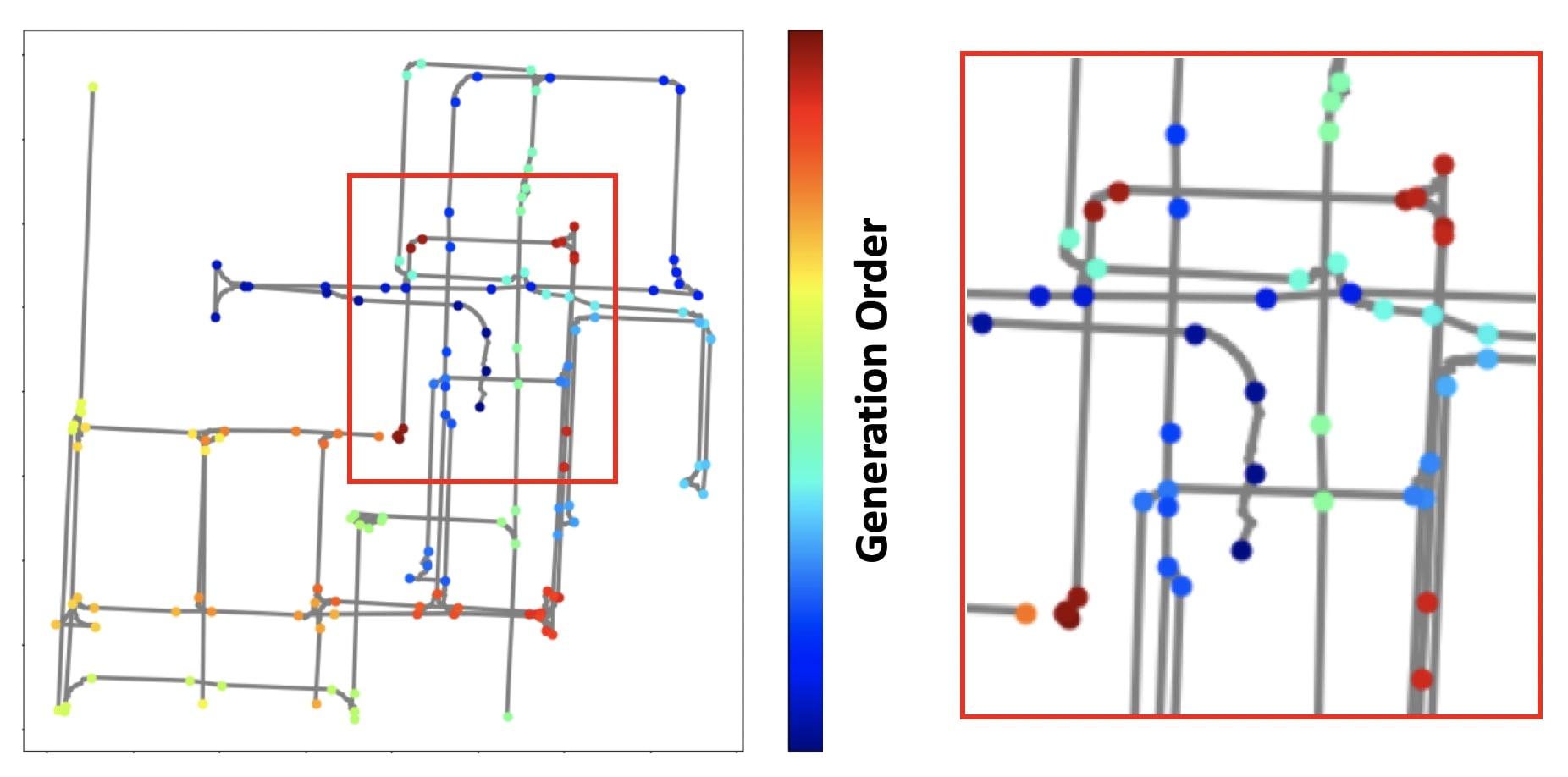}
\end{center}
\vskip -0.5cm
\caption{\lu{The global node generation order (from blue to red) and the scalability of our hierarchical graph generative model HDMapGen to generate HD maps with a FoV of $400m \times 400m$.}}
\vskip -0.5cm
\label{fig: scalability}
\end{figure}

\lu{We further experiment \HDMapGen with a FoV of $400m \times 400m$. As shown in Figure~\ref{fig: scalability}, \HDMapGen consistently achieves promising results for large graph generation, demonstrating its scalability.}

\subsection{Latency Analysis}

We show a latency comparison in Table~\ref{tab: performance}. \HDMapGen achieves a speed-up of ten-fold compared to SketchRNN. \HDMapGen is much more efficient due to (1) using a smaller number of nodes to represent a global graph, and (2) utilizing GRAN with state-of-the-art time-efficiency for global graph generation and a single-shot decoder for the local graph generation. 

\begin{table}[h]
\captionsetup{font=small}
\small
\vspace{-0.2cm}
\setlength{\tabcolsep}{6pt}
\begin{center}
\begin{tabular}{ c c c c }
 \toprule
 Model & \HDMapGen &PlainGen & SketchRNN \\
 \hline
 Time [s] &\textbf{0.20} &0.89 &2.28\\ 
%Complexity &\textbf{$O(N)$} &$O(N^2M)$ &$O(N^2M)$ \\
\bottomrule
\end{tabular}
\end{center}
\vspace{-0.5cm}
\caption{The generation time of an HD map with a FoV of $200m \times
200m$ on Argoverse dataset using \HDMapGen, PlainGen, and SketchRNN. \HDMapGen is clearly the fastest. \vspace{-0.4cm}}
\label{tab: performance}
\end{table}

%GRAN has the  with a complexity of $O(N)$. It outperforms common recurrent models, typically with a complexity of $O(N^2)$.

%will move this to the method part

%\subsection{Comparison of FLOPs and Model Sizes}

\section{Conclusion}
In this work, we introduce a novel and challenging task to generate HD maps in a data-driven way. We performed a systematic exploration for autoregressive generative models with different data representations. Our proposed hierarchical graph generative model \HDMapGen largely outperforms other baselines. We further demonstrated the advantages of \HDMapGen in generation quality, diversity, scalability, and efficiency on real-world datasets. 
%Meanwhile, our proposed hierarchical graph generative model is a principled tool, which has great potentials for other applications.
\section{Acknowledgement}
We would thank Benjamin Sapp and Tianxing He for insightful comments and suggestions.

%%%%%%%%% BODY TEXT

{\small
\bibliographystyle{ieee_fullname}
\bibliography{egbib}
}

\clearpage
\appendix

In the supplementary materials, we describe the statistics of HD map datasets in Section~\ref{supp_sec: data}; more qualitative results of generated maps in Section~\ref{supp_sec: result}; and details of \HDMapGen models in Section~\ref{supp_sec: model}.

\begin{table*}[h]
\captionsetup{font=small}
\small
\begin{center}
\begin{tabular}{ c| c |c c  c c c |c c c c c|c}
 \toprule
 
\multicolumn{2}{c|}{Graph Type} &\multicolumn{5}{|c|}{Plain Graph} & \multicolumn{5}{c}{Global Graph} &\multicolumn{1}{|c}{Local Graph}\\
  \hline
\multicolumn{2}{c|}{Component} &\multicolumn{2}{c|}{\#Nodes} & \multicolumn{3}{c|}{\#Edges} & \multicolumn{2}{c|}{\#Nodes} & \multicolumn{3}{c}{\#Edges} &\multicolumn{1}{|c}{\#Nodes} \\
 \hline
Dataset &City &Max &\multicolumn{1}{c|}{Mean} &Max &Mean &No edge/Edge &Max &\multicolumn{1}{c|}{Mean} &Max &Mean &No edge/Edge &Max\\
\hline
Argoverse &MIA &250 &147 &267 &151 &164 &112 &43 &138 &50 &40 &8\\ 
Argoverse &PIT &250 &178 &265 &185 &187 &111 &51 &134 &64 &44 &8\\ 
In-house &SF &498 &164 &537 &166 &188 &100 &47 &135 &52 &48 &20\\ 
%Complexity &\textbf{$O(N)$} &$O(N^2M)$ &$O(N^2M)$ \\
\bottomrule
\end{tabular}
\end{center}
\caption{Graph statistic for Argoverse Dataset (Miami and Pittsburg) and in-house Dataset (San Francisco): number of nodes, number of edges, the ratio of no edges v.s. edges (sparsity level) in a plain graph or a hierarchical graph (including both global graph and local graph).}
\label{tab: Graph Statistic}
\end{table*}

\section{Dataset Statistics}
\label{supp_sec: data}
In this section, we introduce the statistics of the HD map datasets in Table~\ref{tab: Graph Statistic}. In the plain graph setting, we have a maximum of 250 nodes (control points) for the Argoverse dataset and 498 nodes for our In-house dataset. After representing the map data as hierarchical graphs, we convert 30\% of the original control points into global graph nodes and 70\%  of which into local graph nodes.
For the local graph generation, as we allow a variable number of local nodes in each lane, we set a maximum length of $W$ with a node validity mask. $W$ is set to 8 for the Argoverse dataset and 20 for the In-house dataset.

\section{Qualitative Results}
\label{supp_sec: result}
\subsection{Global Graph Diversity}
\begin{figure*}[h]
\captionsetup{font=small}
\begin{center}
% %\vspace{-40pt}
\includegraphics[width=1\linewidth]{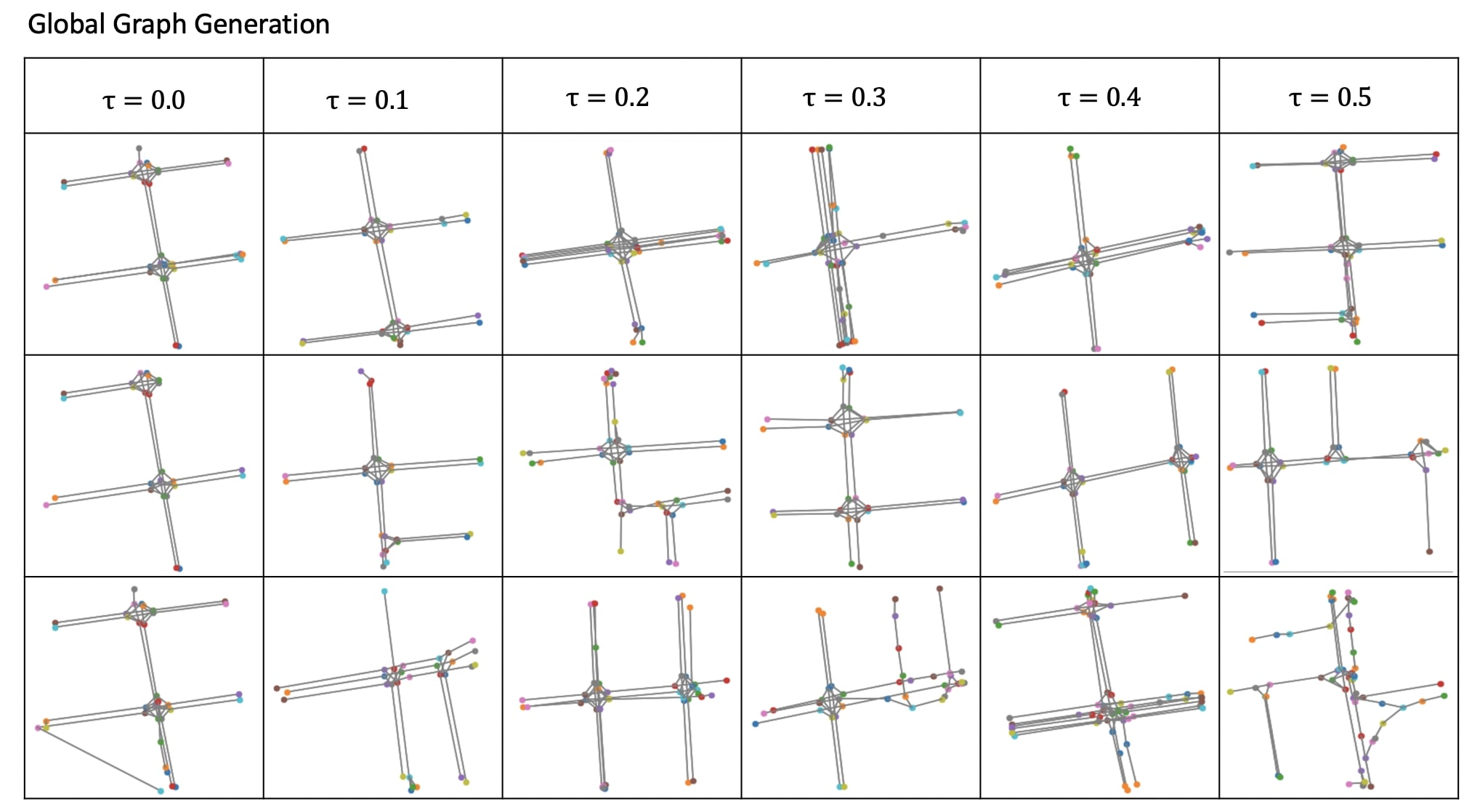}
 \end{center}
\caption{Diversity control for global graph generated from \HDMapGen. Outputs generated with different temperatures $\tau$. }
%The corresponding diversity scores quantified by Chamfer distance are shown in Table~\ref{tab: diversity: chamfer}. }
 \label{fig: global_diversity_supp}
 %\vskip -0.3cm
 \end{figure*}

In Figure~\ref{fig: global_diversity_supp}, we demonstrate more generated global graph results as the temperature changes during the diversity control. We show that more novel HD map patterns (three intersections or more parallel lanes) are generated when a large temperature $\tau$ is applied, yet the quality-diversity trade-off still exists.
%, the more novelty includes while the fidelity then might suffer. 

%Meanwhile, we also use a metric quantified by Chamfer distance to evaluate the diversity of generated global graph. We evaluate the diversity on two levels, one is the novelty compare to all ground-truth maps. And another one is internal diversity among all outputs. For each map, we represent each lane as a data point with $4$-dimensional features -- the $x$ and $y$ coordinates of two end points of each lane. And then we compare each generated map with all maps in the target set with Chamfer distance. Then we calculate the average lowest values for all generated maps as the diversity score. 

 \subsection{HDMapGen Generation Quality}

\begin{figure*}[t]
\captionsetup{font=small}
\begin{center}
% %\vspace{-40pt}
\includegraphics[width=1\linewidth]{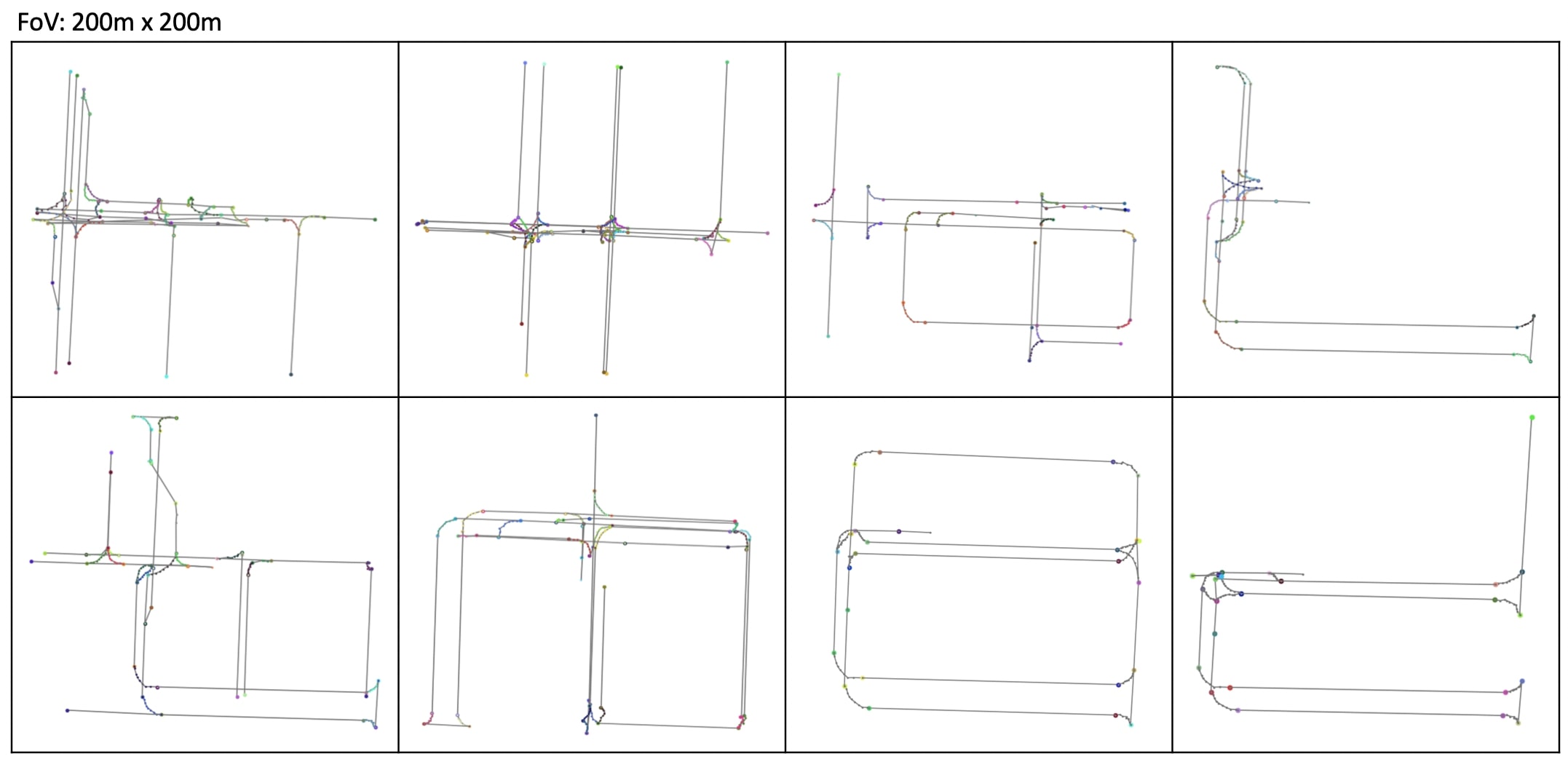}
 \end{center}
\caption{Hierarchical graph with a FoV of $200m \times 200m $ generated from \HDMapGen.}
 \label{fig: 200m maps}
 %\vskip -0.3cm
 \end{figure*}

\begin{figure*}[t]
\captionsetup{font=small}
\begin{center}
% %\vspace{-40pt}
\includegraphics[width=1\linewidth]{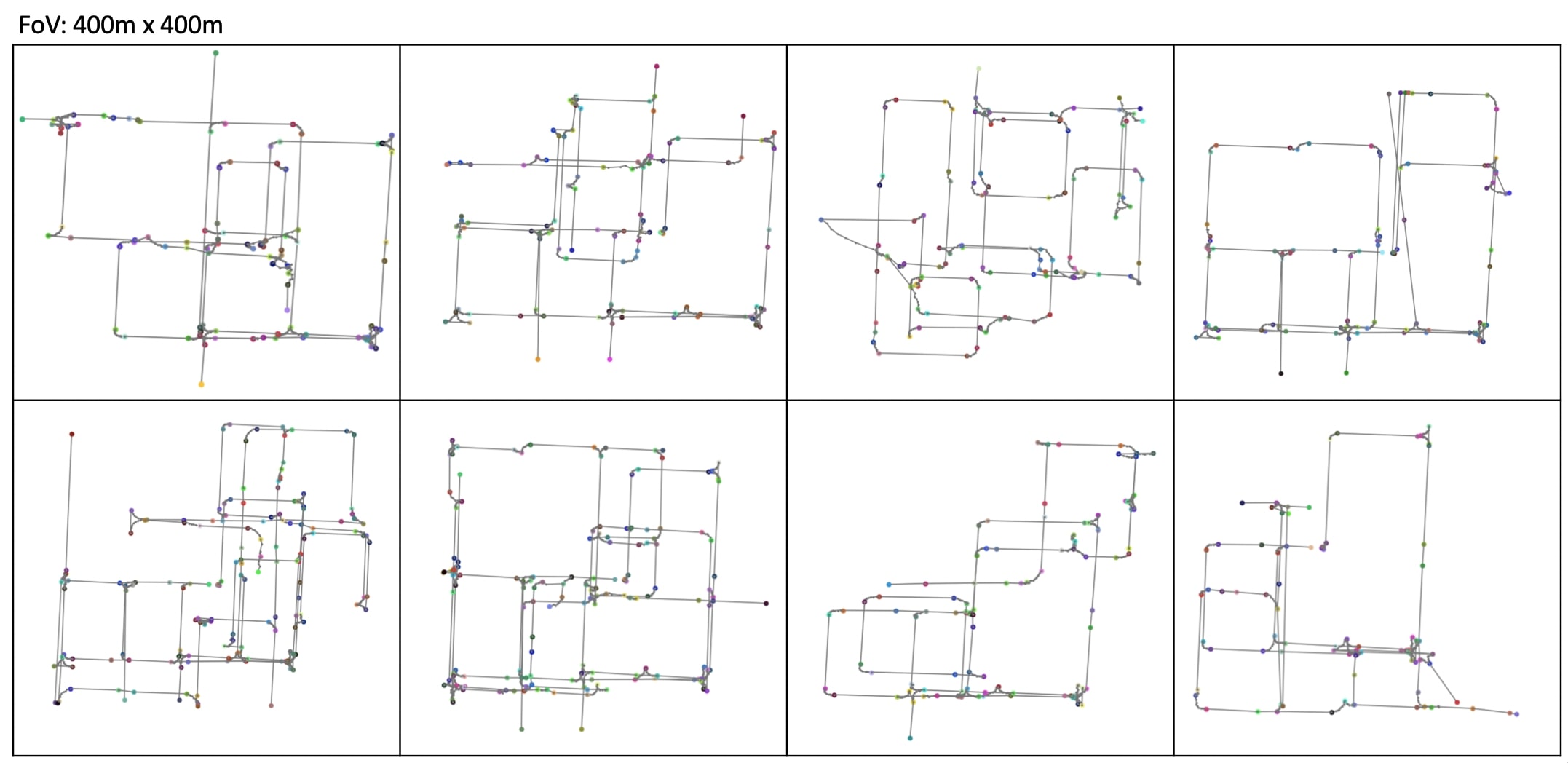}
 \end{center}
\caption{Hierarchical graph with a FoV of $400m \times 400m $ generated from \HDMapGen.}
 \label{fig: 400m maps}
 %\vskip -0.3cm
 \end{figure*}

We show more results from \HDMapGen, generated maps with a FoV of $200m \times 200m $ in Figure~\ref{fig: 200m maps}, and generated maps with a FoV of $400m \times 400m $ in Figure~\ref{fig: 400m maps}.  The average number of global nodes in the smaller maps is $43$, and the average number of global edges is $50$, while for the larger maps, the average number of global nodes is $136$, and the average number of global edges is $175$, which means more than four times of nodes. While our proposed \HDMapGen still achieves promising results for large graph generation, which demonstrates its large scalability.

Notice that an issue with our current results is the unsmoothness of the local graph in some generated samples. However, all of our demonstrated results are not post-processed or applied with any heuristic thresholding or filtering during generation. One can expect a better version with additional steps as described above.

 \section{Model Details}
\label{supp_sec: model}
In this section, we introduce more details of baseline and two variants of \HDMapGen models.

\subsection{Global Graph Generation: Two Variants}
\begin{figure*}[t]
\captionsetup{font=small}
\begin{center}
\vspace{-10pt}
\includegraphics[width=0.8\linewidth]{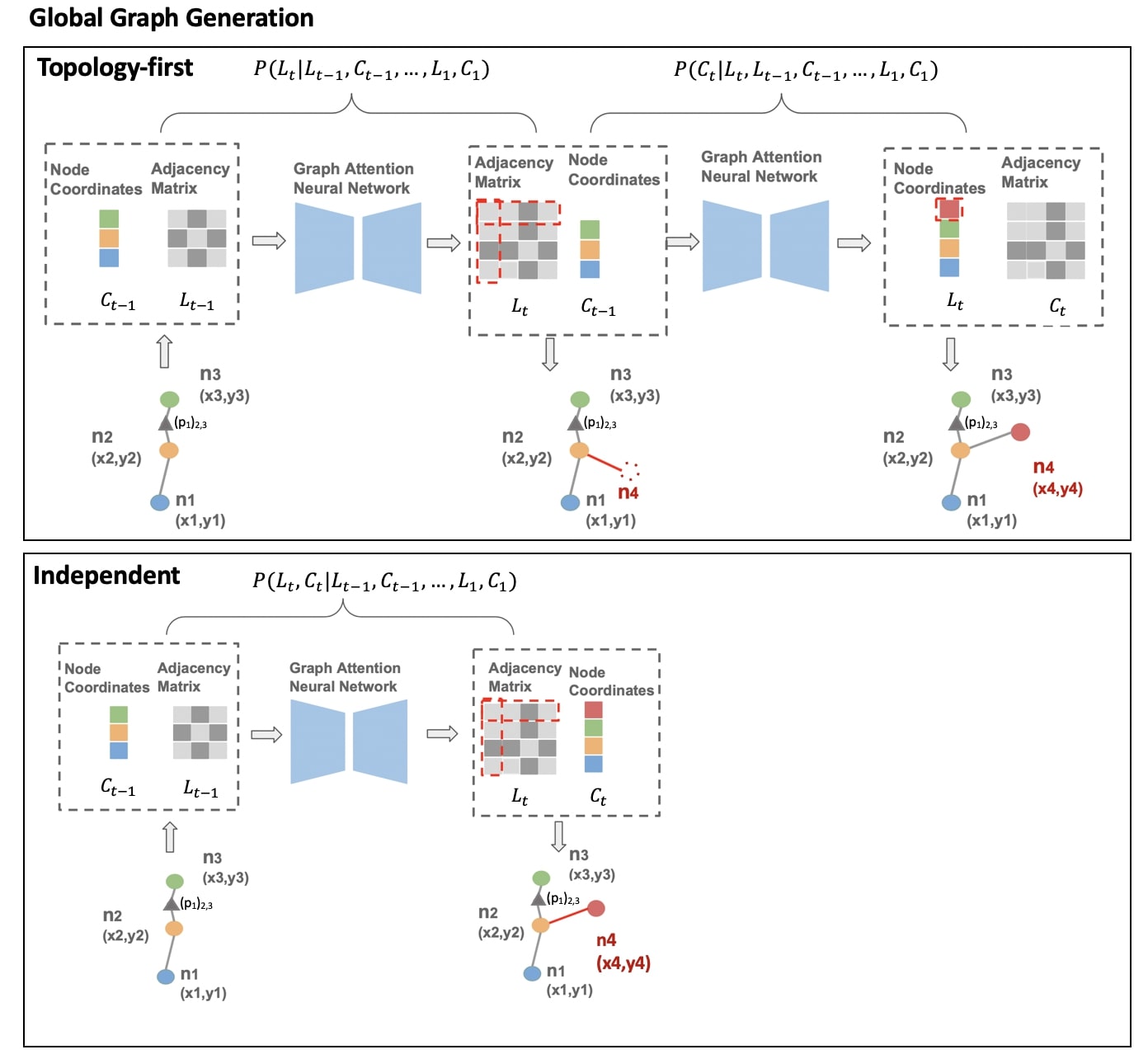}
 \end{center}
\caption{Two variants of \HDMapGen: \textit{topology-first} and \textit{independent}. The models consider different dependence and priority to generate coordinates and topology at each step. 
%global graph generator (top) produces a new node with its coordinates and its connections to existing nodes, which is a row and a column in the adjacency matrix
\vspace{-0.6cm}}
 \label{fig: Topology-first and Independent}
 %\vskip -0.3cm
 \end{figure*}

In Figure~\ref{fig: Topology-first and Independent}, we show the neural network structures of the other two variants of \HDMapGen using \textit{topology-first} or \textit{independent} for global graph generation. 

\noindent\textbf{\textit{Topology-first}}: At the step $t$, the model firstly generated the topology $L_t$ to generate the connections of node $t$ with previously generated nodes. And then taking the new topology information $L_t$ as additional inputs to the GRAN model to generate the spatial coordinates $C_t$ of node $t$.

\noindent\textbf{\textit{Independent}}: At the step $t$, the model is trained to generate $C_t$ and $L_t$ simultaneously, while not consider any dependence between these two variables. 

% \subsection{Local Graph Generation: Number of Nodes Variations}

% The local graph to reveal the details of each global edge is generated after every node $t$ and corresponding new edges $L_t$ is generated. As the local graph always has a unique topology as a sequence of control points, we model it as a sequence of coordinate mask $\{(C_j)_{t,s}\}$ and a corresponding valid mask $\{(M_j)_{t,s}\}$. The valid mask enables a variation in the number of nodes in each graph. For the straight lines which have filtered all redundant control points after map preprocessing, the valid mask $\{(M_j)_{t,s}\}$ is all $0s$, while for the lanes at the corner which usually have multiple control points remained to guarantee the smoothness, $\{(M_j)_{t,s}\}$ is likely to have more values of $1$.

\subsection{Baselines}
In this work, we implement two baselines SketchRNN~\cite{ha2017neural} and PlainGen. For SketchRNN which uses a sequence generative model, we use ``layer norm" model as our encoder and decoder model. We also apply different temperatures to fine-tune a better output. For a plain graph generative model, we use PlainGen, a model derived from our global graph generation step of \HDMapGen. Notice that NTG ~\cite{chu2019neural} which is designed for a more coarse road layout generation, also uses a plain graph generative model. However, since the model has not been open-sourced yet, we are not able to perform a comparison with NTG on this high-definition map generation task.

\end{document}